\documentclass[10pt,twocolumn,letterpaper]{article}
\usepackage[table]{xcolor}
\usepackage[pagenumbers]{cvpr}
\definecolor{cvprblue}{rgb}{0.21,0.49,0.74}
\usepackage[pagebackref,breaklinks,colorlinks,allcolors=cvprblue]{hyperref}
\usepackage{bbm}
\usepackage{makecell}
\usepackage{booktabs}
\usepackage{multirow}
\usepackage{tikz}

\newcolumntype{M}{>{\centering\arraybackslash}m{1.2cm}}

\title{Enhancing Diffusion-based Unrestricted Adversarial Attacks via Adversary Preferences Alignment}

\author{
Kaixun Jiang\textsuperscript{1},  
Zhaoyu Chen\textsuperscript{1},  
Haijing Guo\textsuperscript{2}, 
Jinglun Li\textsuperscript{1},
Jiyuan Fu\textsuperscript{2},  \\
Pinxue Guo\textsuperscript{1}, 
Hao Tang\textsuperscript{3}, 
Bo Li\textsuperscript{4}, 
Wenqiang Zhang\textsuperscript{1,2}  \\[0.5em] 
\textsuperscript{1} College of Intelligent Robotics and Advanced Manufacturing, Fudan University \\
\textsuperscript{2} Shanghai Key Lab of Intelligent Information Processing,\\ College of Computer Science and Artificial Intelligence, Fudan University\\ 
\textsuperscript{3}Peking University \ \ \ \
\textsuperscript{4} vivo Mobile Communication Co., Ltd \\ 
}

\begin{document}
\maketitle
\renewcommand{\thefootnote}{\fnsymbol{footnote}}  
\begin{abstract}
 Preference alignment in diffusion models has primarily focused on benign human preferences (e.g., aesthetic). In this paper, we propose a novel perspective: framing unrestricted adversarial example generation as a problem of aligning with adversary preferences. Unlike benign alignment, adversarial alignment involves two inherently conflicting preferences: visual consistency and attack effectiveness, which often lead to unstable optimization and reward hacking (e.g., reducing visual quality to improve attack success). To address this, we propose APA (Adversary Preferences Alignment), a two-stage framework that decouples conflicting preferences and optimizes each with differentiable rewards. In the first stage, APA fine-tunes LoRA to improve visual consistency using rule-based similarity reward. In the second stage, APA updates either the image latent or prompt embedding based on feedback from a substitute classifier, guided by trajectory-level and step-wise rewards. To enhance black-box transferability, we further incorporate a diffusion augmentation strategy. Experiments demonstrate that APA achieves significantly better attack transferability while maintaining high visual consistency, inspiring further research to approach adversarial attacks from an alignment perspective. Code will be available at https://github.com/deep-kaixun/APA.
\end{abstract}

\section{Introduction}
% \vspace{-10pt}
\begin{figure}
  \centering
  % \vspace{-25pt}
  \includegraphics[width=0.48\textwidth]{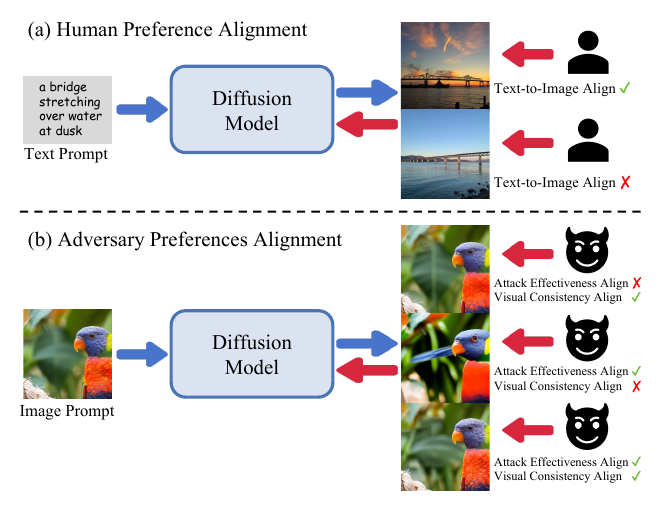}
   % \vspace{-10pt}
  \caption{Comparison of Human Preference Alignment and Adversary Preferences Alignment.}
  % \vspace{-10pt}
  \label{fig:intro}
\end{figure}

Preference alignment, adapting pre-trained diffusion models~\cite{ldm} for diverse human preferences, is increasingly prominent in image generation. This typically involves modeling human preferences with explicit reward models or pairwise data~\cite{dpo}, then updating model policies via reinforcement learning~\cite{ddpo,dpok} or backpropagation with differential reward~\cite{draft,alignprop,textcraftor}. However, current research largely centers on benign human preferences like aesthetics and text-image alignment (Figure~\ref{fig:intro}(a)), malicious adversary preferences alignment, where security researchers use diffusion models to create unrestricted adversarial examples~\cite{aca} has received limited attention. These examples are vital for assessing the adversarial robustness~\cite{aca,jiang2023efficient,jiang2023exploring,jiang2023towards} of deep learning models. Adversaries, as depicted in Figure~\ref{fig:intro}(b), primarily seek two preferences: 1) Visual consistency: Ensuring that generated images have minimal, semantically negligible differences from the original. 2) Attack effectiveness: Achieving high transferable attack performance, where adversarial examples generated from a surrogate model fool black-box target models~\cite{mi,difgsm}.

Aligning with such adversarial preferences presents two major challenges. First, preference data is unavailable. Existing diffusion-based attacks~\cite{aca,diffattack} build on the idea of traditional $L_p$ attacks~\cite{mi,pgd}, adapting the strategy of adding optimized perturbations from the pixel space to the latent space. However, latent spaces in diffusion models are highly sensitive—even slight perturbations can result in severe semantic drift. This makes it infeasible to obtain stable, preference-consistent adversarial examples for pairwise data collection, rendering traditional preference optimization techniques like DPO~\cite{dpo} or unified reward modeling inapplicable.
Second, these preferences are inherently in conflict. Joint optimization with reward weighting often results in reward hacking~\cite{geirhos2020shortcut}, (e.g., one shortcut to improving attack success is to reduce visual consistency), leading to unstable or degenerate solutions (Figure~\ref{fig:aba}(b)).

To address this, we introduce APA (Adversary Preferences Alignment), a novel two-stage framework that separates and sequentially optimizes the adversary preferences using direct backpropagation with differentiable rewards. Specifically: 1) Visual Consistency Alignment: We use a differentiable visual similarity metric as a rule-based reward and perform policy updates by fine-tuning the diffusion model’s Low-Rank Adaptation (LoRA) parameters~\cite{hu2022lora}. This stage encodes the input image’s structure into the model’s generation space, forming a visually stable foundation for downstream attack optimization. 2) Attack Effectiveness Alignment: We optimize either the image latent or the prompt embedding based on feedback from a white-box surrogate classifier. This process uses dual-path attack guidance (both trajectory-level and step-wise dense rewards) to align with the adversary’s attack preference. To prevent overfitting to the surrogate, we introduce a diffusion augmentation strategy that aggregates gradients from intermediate steps to increase diversity, thereby improving black-box transferability. Our framework explicitly decouples these conflicting preferences to mitigate reward hacking, enabling more controllable and scalable adversary preferences alignment. Our contributions are summarized as follows:

$\quad\bullet$ We are the first to transform unrestricted adversarial attacks into adversary preferences alignment (APA) and propose an effective two-stage APA framework.

$\quad\bullet$ Our APA framework decouples adversary preferences into two sequential stages which include LoRA-based visual consistency alignment using a rule-based visual similarity reward and attack effectiveness alignment guided by dual-path attack guidance and diffusion augmentation.

$\quad\bullet$ APA achieves state-of-the-art transferability against both standard and defense-equipped models while preserving high visual consistency. Our framework is flexible and scalable, supporting various diffusion models, optimization parameters, update strategies, and downstream tasks.

\section{Related Work}
\vspace{-10pt}
\noindent \textbf{Unrestricted Adversarial Attacks.} Unrestricted adversarial attacks address key limitations of traditional $L_p$ attacks, which apply pixel-level perturbations that are often perceptible due to distribution shifts from clean images~\cite{DBLP:conf/eccv/JohnsonAF16} and are increasingly countered by existing defenses~\cite{diffpure,singh2023revisiting}. Instead, unrestricted attacks generate more natural examples by subtly modifying the semantic content of the original images. Early approaches focused on single-type semantic perturbations, including shape~\cite{stadv,ADer}, texture~\cite{semanticadv,LPA}, and color~\cite{aca} manipulations. Shape-based methods use deformation fields to induce structural changes; texture-based methods modify image texture or style—for example, DiffPGD~\cite{diffpgd} adds $L_p$ perturbations in pixel space followed by diffusion-based translation, thus falling into this category. Color-based attacks (e.g., SAE~\cite{sae}, ReColorAdv~\cite{ReColorAdv}, ACE~\cite{ACE}) adjust hue, saturation, or channels to improve visual naturalness, often at the cost of transferability. However, these methods typically optimize a single semantic factor, limiting their generality and expressiveness. Recent efforts leverage the latent space of generative models to produce more flexible adversarial examples. In particular, diffusion models have been adapted for this purpose~\cite{aca, diffattack, pan2024sca, dai2025advdiff, chen2023advdiffuser}. For instance, ACA~\cite{aca} employs DDIM inversion and skip gradients to optimize latent representations. Nonetheless, due to the sensitivity of latent space manipulations, existing approaches often struggle to preserve the visual semantics of the original input. We address this by reframing adversarial attack generation as a preference alignment problem, and propose a two-stage framework that decouples conflicting visual consistency and attack effectiveness for more stable generation.

\noindent \textbf{Alignment of Diffusion Models.} Preference alignment for diffusion models aims to optimize the pretrained diffusion model based on human reward.  Current methods adapted from large language models (LLMs) include reinforcement learning (RL)~\cite{ddpo, dpok, rl2,lee2024parrot}, direct preference optimization (DPO)~\cite{dpo}, and direct backpropagation using differentiable rewards (DR)~\cite{draft, alignprop, textcraftor}. RL approaches frame the denoising process as a multi-step decision-making process, often using proximal policy optimization (PPO)\cite{ppo} for fine-tuning. Although flexible, RL is inefficient and unstable when managing conflicting rewards like visual consistency and attack effectiveness. DPO, which avoids reward models by ranking outputs with the Bradley-Terry model, faces challenges in adapting to adversarial preference alignment (APA), given the difficulty of obtaining high-quality adversarial examples for ranking. DR is efficient, using gradient-based optimization with differentiable rewards. To ensure flexibility and stability, we propose a two-stage framework built on DR. By decoupling conflicting objectives into differentiable rewards, it effectively addresses the challenges of multi-preferences alignment.

\section{Method}
\subsection{Preliminary}
\vspace{-5pt}
\noindent \textbf{Unrestricted Adversarial Example (UAE).} 
Given a clean image $x$, considering both visual consistency and attack effectiveness, the optimization objective for UAE $x_{adv}$ can be expressed as:
% \scalebox{0.}{}
\begin{equation}
    \underset{x_{adv}}{\max} \ f_{\phi^{'}}(x_{adv}) \neq y,  s.t.\ x_{adv} ~\mathrm{is~naturally~similar~to}~x ,
    \label{eq:attack_definition_ori}
\end{equation}
where $y$  denotes the label of $x$, and  $f_{\phi^{{\prime}}}(\cdot)$  represents target models for which gradients are inaccessible for direct optimization. Since natural similarity cannot be enforced via $L_p$ norms as in perturbation-based attacks~\cite{pgd,mi}, unrestricted adversarial attacks must search for optimal adversarial examples in both conflicting optimization spaces.

\noindent \textbf{Latent Diffusion Model (LDM).} LDM~\cite{ldm} is a latent variable generative model trained on large-scale image-text pairs, relying on an iterative denoising mechanism. During training, the denoising model $\epsilon_\theta(\cdot)$ is trained by minimizing the variational lower bound loss function, typically using a UNet to predict the noise added to the original data:
\begin{equation}
   \underset{\epsilon_\theta}{\min} \quad E_{t \sim[1, T], \epsilon \sim \mathcal{N}(0, \mathbf{I})}\left\|\epsilon-\epsilon_\theta\left(z_t, t,c\right)\right\|^2,
   \label{eq:train}
\end{equation}
where $t$ denotes the timestep, $T$ denotes the total number of timesteps, and  $\epsilon$ is the random noise sampled from $\mathcal{N}(0,\mathbf{I})$. The latent variable  $z_t$  is generated by adding noise to  $z_0$  over  $t$  steps, where $z_0$ is the latent representation of the original input $x$ obtained through encoder $\mathcal{E}(\cdot)$, i.e.,  $\mathcal{E}(x)=z_0$. The diffusion process is defined as  $q(z_t|z_0) = \sqrt{\bar{\alpha}_{t}} \cdot {z}_0+\sqrt{1-\bar{\alpha}_{t}} \cdot \epsilon$, where $\sqrt{\bar{\alpha}_{t}}$  is a hyperparameter that controls the level of noise added at each timestep $t$~\cite{ddpm}.  $c$ denotes the conditioning text. During inference, we typically sample  $z_T$  from  $\mathcal{N}(0, \mathbf{I})$ , and then use the DDIM denoising~\cite{ddim} to iteratively denoise $z_T$. The iterative denoising process can be expressed as:
\begin{equation}
\scalebox{0.85}{$
% \begin{aligned}
{z}_{t-1}=\sqrt{\bar{\alpha}_{t-1}}\left(\frac{{z}_t-\sqrt{1-\bar{\alpha_t}} \epsilon_\theta\left({z}_t, t, c\right)}{\sqrt{\bar{\alpha_t}}}\right)
        +\sqrt{1-\bar{\alpha}_{t-1}} \epsilon_\theta\left({z}_t, t, c\right).$}
\label{eq:ddim}
% \end{aligned}
\end{equation}
After $T$  steps of DDIM denoising, the resulting  $\bar{z}_0$  is decoded into pixel space via a decoder $\mathcal{D}(\cdot)$, generating an image that matches the condition $c$. For tasks with a given reference image (e.g. image editing),  $z_T$ is typically not sampled from random noise, instead, it is obtained through DDIM Inversion~\cite{ddim} based on the reference image $x$:
\begin{equation}
\scalebox{0.8}{$
% \begin{aligned}
{z}_{t} = \sqrt{\bar{\alpha_{t}}} \left(\frac{{z}_{t-1} - \sqrt{1 - \bar{\alpha}_{t-1}} \, \epsilon_\theta\left({z}_{t-1}, t, c\right)}{\sqrt{\bar{\alpha}_{t-1}}}\right) 
          + \sqrt{1 - \bar{\alpha}_{t}} \, \epsilon_\theta\left({z}_{t-1}, t, c\right),
% \end{aligned}
$}
\label{eq:ddim_inversion}
\end{equation}
where initial $z_0$ denotes the latent of reference image $x$. Through whole DDIM inversion, i.e., iteratively using Eq.~\ref{eq:ddim_inversion} $T$ times, we obtain $z_T$, which preserves information of $x$.  

% \section{Method}
\subsection{Adversary Preferences Alignment Framework}
% \vspace{-10pt}
Existing works~\cite{aca} leverage the natural image generation capabilities of diffusion models to generate unrestricted adversarial examples, the adversary optimizes $z_T$ (obtained via DDIM Inversion) instead of directly optimizing $x$ in the pixel space. The mapping from $z_T$ to $x_{{adv}}$ is defined as
$x_{\text{adv}} = \mathcal{D}(\bar{z}_0)$, where $\bar{z}_0$ is obtained by applying $T$ steps of DDIM denoising: $\bar{z}_0 = \underbrace{{De} \circ \cdots \circ {De}}_{T}(z_T)$. This sequential process defines a denoising trajectory. However, these methods attempt to solve Eq.~\ref{eq:attack_definition_ori} by jointly optimizing $z_T$. Due to the sensitivity of the latent space to noise (shown in Figure~\ref{fig:m1}) and the mutual exclusivity of the two optimization objectives (shown in Figure~\ref{fig:aba}(b)), the generated adversarial examples often fall into a suboptimal trade-off between the conflicting objectives.

\begin{figure*}[!t]
    \centering
    \scalebox{0.8}{
    \includegraphics{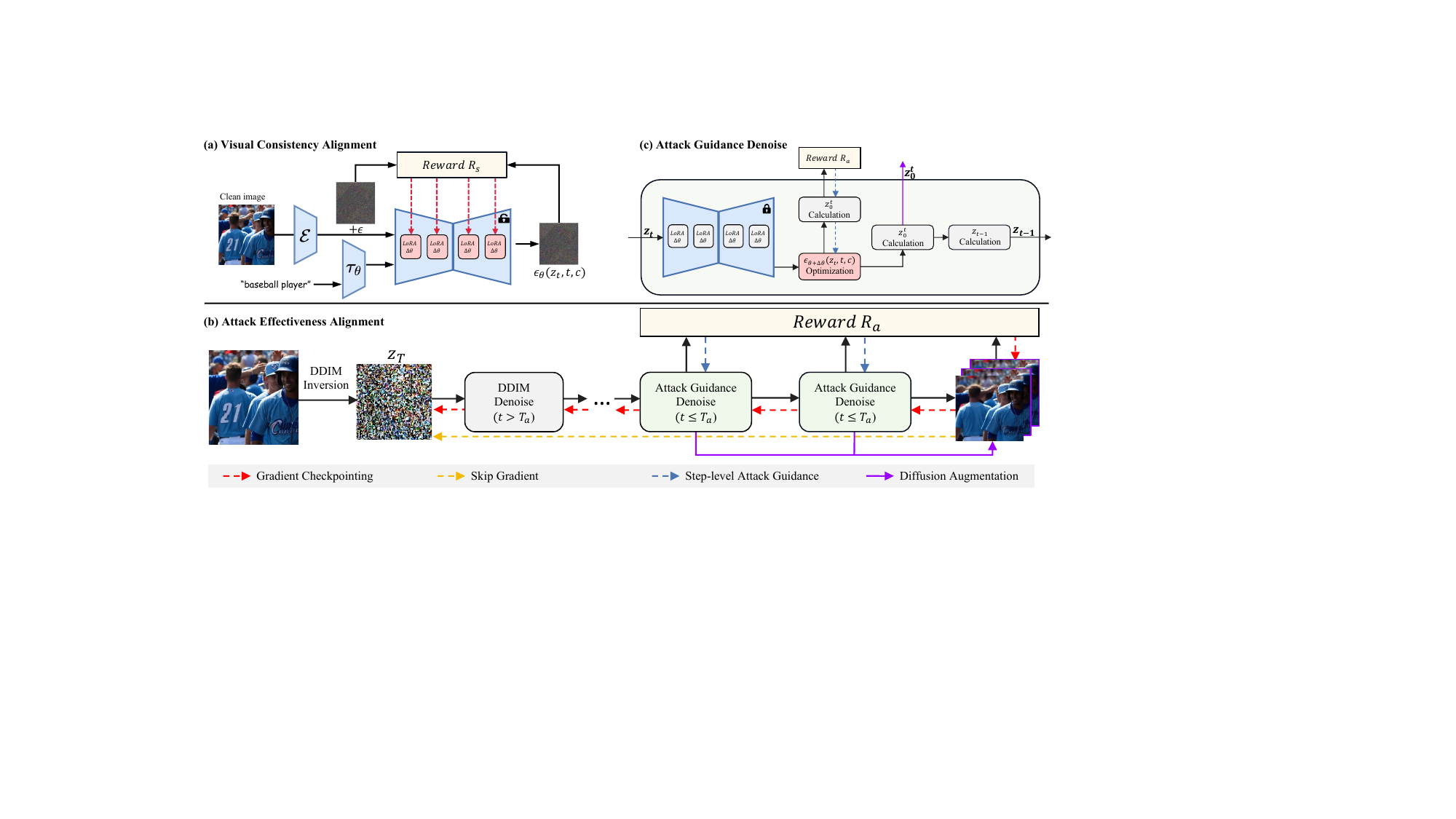}}
    \vspace{-5pt}
    \caption{Overview of our APA framework. APA first optimizes the LoRA parameters with a visual consistency reward, storing input image information in LoRA. Then, the input image undergoes DDIM inversion to obtain $z_T$. After DDIM denoising and attack guidance denoising, APA generates trajectory-level $\bar{z}_0$  and diffusion augmentation output  $z^t_{0}$,  mixing them using Eq.~\ref{eq:mix} and passing to the substitute classifier to calculate $R_a$. Finally,  $z_T$ is iteratively optimized using skip gradient (APA-SG) or gradient checkpointing (APA-GC).}
    \vspace{-10pt}
    \label{fig:main}
\end{figure*}

To address this, we propose a two-stage adversary preferences alignment framework (APA): first, we reframe unrestricted adversarial attacks as a multi-preference alignment problem and decouple visual consistency and attack effectiveness to independent reward models. Then we strengthen visual consistency via LoRA fine-tuning in the first stage and focus on attack effectiveness through dual-path attack guidance and diffusion augmentation in the second stage. Our APA separates visual consistency and attack effectiveness by independently modeling and aligning preference rewards. It then maximizes attack performance within the optimal solution space of visual consistency, achieving closer Pareto optimality. Figure~\ref{fig:main} presents the overall framework of APA.
 
\subsection{Visual Consistency Alignment}
\begin{figure}
  \centering
  % \vspace{-15pt}
  \includegraphics[width=0.48\textwidth]{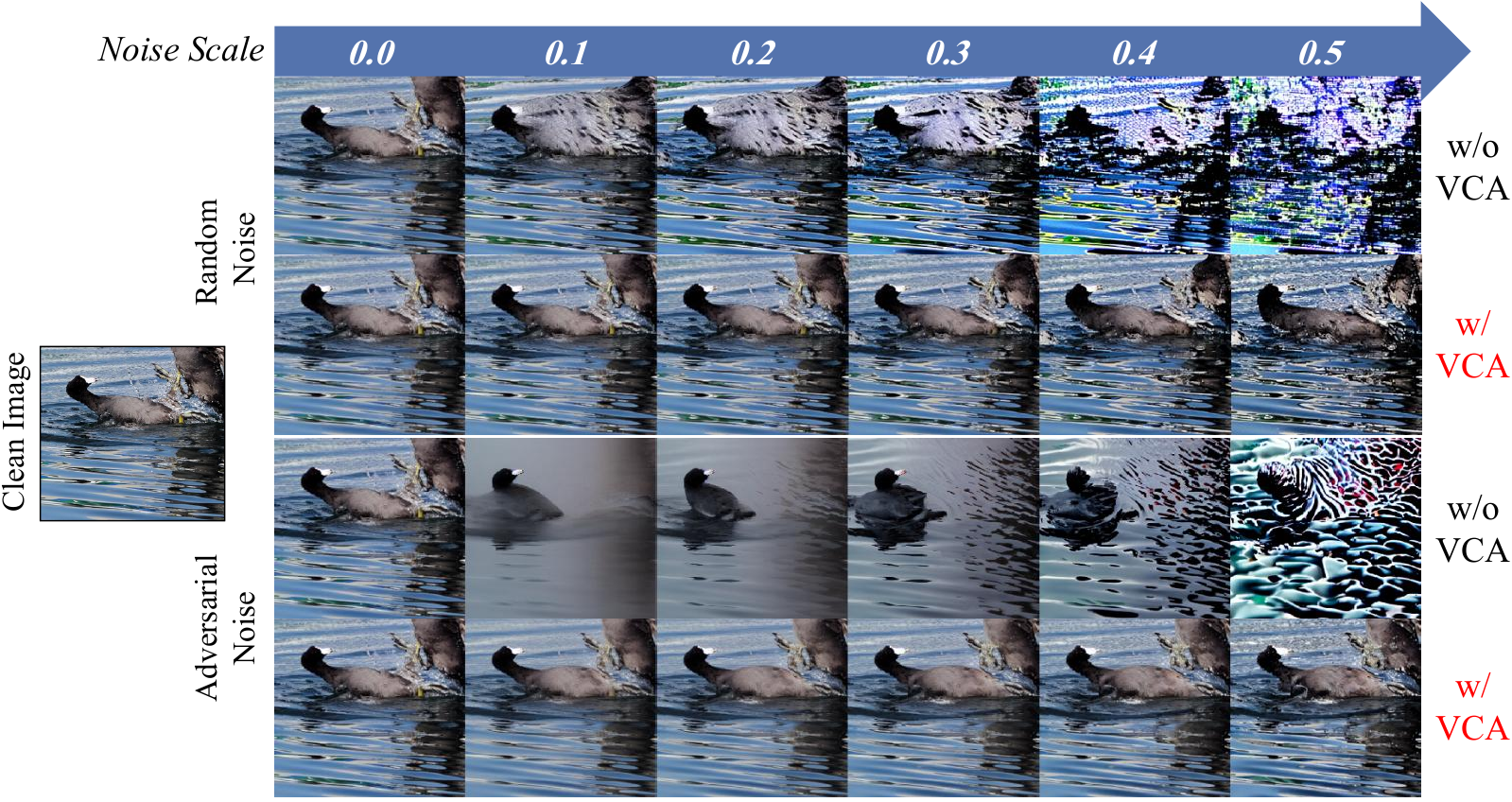}
  % \vspace{-15pt}
  \caption{Impact of adversarial and random noise on $z_T$ in generated images. VCA denotes visual consistency alignment. Our LoRA-based VCA, demonstrates improved noise robustness.}
  \vspace{-5pt}
  \label{fig:m1}
\end{figure}
Diffusion models derive their capabilities from training on extensive images, meaning that even minor changes to the latent or prompt can result in substantially different generations. As shown in Figure~\ref{fig:m1}, without perturbations to the $z_T$ obtained via DDIM Inversion, the model nearly reconstructs the clean image after $T$ denoising steps. However, minor noise, particularly adversarial noise, can cause the generated image to lose visual consistency with the original. To preserve visual consistency during adversarial optimization, we aim to strengthen the diffusion model’s retention of the input image $x$. A straightforward approach is to fine-tune the UNet to overfit the input image, but this risks catastrophic forgetting, degrades image quality, and is inefficient. Previous research in customized generation~\cite{gu2024mix,zhang2024diffmorpher} suggests that LoRA~\cite{hu2022lora} efficiently encodes high-dimensional image semantics into the low-rank parameter space. Thus, we adopt LoRA $\Delta \theta$  as the policy model during the visual consistency alignment stage. Then, we need to determine a reward model or reward function $R_s(\cdot)$ to optimize $\Delta \theta$. The straightforward approach is to compute the visual similarity $S(\cdot)$ between the original input and the output of the diffusion model, as follows:
\begin{equation}
    % \vspace{-5pt}
    \underset{\Delta \theta}{\max}~R_s(\Delta \theta) = S(\mathcal{D}(\bar{z}_0),x),
    \label{eq:train_lora1}
    % \vspace{-5pt}
\end{equation}
where $\bar{z}_0$ represents the $T$-step denoised output, requiring $T$ computations of Eq.~\ref{eq:ddim}. To reduce this, we first shift the similarity metric to the latent space, calculating $S(\bar{z}_{0}, z_0)$. We then approximate trajectory-level similarity by accumulating similarity across all steps. Thus, inspired by Eq.~\ref{eq:train}, $R_s(\Delta \theta)$ is reformulated as:
\begin{equation}
% \vspace{-5pt}
R_s(\Delta \theta)=E_{t,\epsilon} - \|\epsilon-\epsilon_{\theta+\Delta \theta}\left(z_t, t,c\right)\|^2,
\label{eq:rs}
\end{equation}
where $t \in [1,T]$. Since Eq.~\ref{eq:rs} is differentiable, we can update $\Delta \theta$ via the direct backpropagation~\cite{draft,alignprop} to maximize the reward, as follows:
$\Delta \theta= \Delta \theta + \alpha \nabla_{\Delta \theta} R_s$, where $\alpha$ represents the learning rate. Finally, $\Delta \theta$ is integrated into $\epsilon_\theta$, enabling the model to generate visually consistent outputs whether regular noise or adversarial noise is applied to $z_T$, as illustrated in Figure~\ref{fig:m1}.

% \vspace{-5pt}
\subsection{Attack Effectiveness Alignment}
In this stage, We use $z_T$ obtained via DDIM inversion as the optimization variable (optional prompt embedding discussed in Section~\ref{apa-p}). Next, we need to model the reward $R_a$ for attack effectiveness alignment. First, using $f_\phi^{'}(\cdot)$ in Eq.~\ref{eq:attack_definition_ori} directly as the reward model results in sparse rewards of only 1 or 0, significantly increasing optimization difficulty. To address this, we draw inspiration from traditional transfer attacks~\cite{tifgsm,mi} and choose a differentiable surrogate model $f_\phi(\cdot)$ as the reward model. This allows optimization via direct backpropagation based on gradients~\cite{draft,alignprop} Additionally, to mitigate the gap between the surrogate model and the target model, we propose diffusion augmentation to enhance generalization and alleviate potential reward hacking~\cite{geirhos2020shortcut}. Thus, the attack effectiveness reward is formulated as: $R_a(z_T) = L(f_\phi(x_{{adv}}), y)$,  $L(\cdot) $ denotes cross-entropy loss.

\noindent \textbf{Dual-path Attack Guidance.} We refer to the generation of $x_{adv}$ through $T$-step denoising from $z_T$ as the generation trajectory. Additionally, to enhance gradient consistency, following ACA~\cite{aca}, we use a momentum-based gradient update to optimize $z_T$, encouraging the model to higher $R_a$ output. Thus, our trajectory-level attack optimization can be expressed as:
\begin{equation}
\begin{aligned}
       & g_{tr} = \nabla_{z_T} R_a(f_\phi(x_{adv}),y), \\
       & m^{i}_{tr}=m^{i-1}_{tr}+\frac{g_{tr}}{\|g_{tr}\|_1},\ z_T = \Pi_{z^0_T+\epsilon_a} (z_T + \mu \cdot \text{sgn}(m^i_{tr})), 
       \label{eq:tr-level}
\end{aligned}
\end{equation}
where  $g_{tr}$  denotes the trajectory-level gradient, $m^i_{tr}$  denotes the momentum of $i^{th}$ trajectory-level attack iteration, $\Pi_{z^0_T+\epsilon_a}$ keeps $z_T$ remain within the $\epsilon_a$-ball centered at the original latent $z^0_T$, $\text{sgn}(\cdot)$ denotes sign function. 

Since solving  $g_{tr}$  requires computing the gradient across the entire trajectory, direct calculation would require extensive memory. One solution is skip gradient~\cite{aca}, which approximates  $g_{tr}$  as  $\rho \cdot \nabla_{\bar{z}_0} R_a(f_\phi(x_{adv}), y)$~\cite{aca}, avoiding memory use for  $T$-step denoising. The other is gradient checkpointing~\cite{gc}, which reduces memory during backpropagation by selectively storing intermediate activations, enabling direct computation of $g_{tr}$.

Both skip gradient and gradient checkpointing focus on optimizing  $z_T$  from a global trajectory level, where each denoising step uses the same  $R_a$. However, different steps contribute uniquely to the final output: larger timesteps affect structure, while smaller ones refine details~\cite{spo}. As a result, using the same  $R_a$  for attack guidance may cause misalignment, reducing attack effectiveness.

To address this issue, we incorporate step-level attack guidance into the denoising steps. Motivated by class-guided generation~\cite{guide-diffusion}, we introduce the attack reward $R_a$  during each denoising step to guide the noise optimization:
\begin{equation}
\scalebox{0.88}{$
    \epsilon_{\theta+\Delta \theta}(z_t,t,c) = \epsilon_{\theta+\Delta \theta}(z_t,t,c)-\sqrt{1-\bar{\alpha_t}}\nabla_{z_t} R_a(f_\phi(\mathcal{D}(z_t)), y).$}
    \label{eq:classgudied}
\end{equation}  
Each denoising step adjusts the generation direction based on the current step’s $R_a(f_\phi(\mathcal{D}(z_t)), y)$, gradually aligning the final image with higher $R_a$.

Since  $z_t$  is an intermediate denoising result, directly inputting it to the classifier biases reward calculation, as classifiers are typically trained on clean samples. A noise-robust classifier could reduce this bias~\cite{guide-diffusion}, but it would raise training costs and may introduce inconsistencies between substitute and target classifiers, affecting attack performance. To address this, we first replace  $z_t$  with the intermediate result $z^t_0$  generated by DDIM, which represents the estimated trajectory-level $\bar{z}_0$  based on the current step:
\begin{equation}
    {z}^t_0=\frac{{z}_t-\sqrt{1-\bar{\alpha_t}} \epsilon_{\theta+\Delta \theta}\left({z}_t, t, c\right)}{\sqrt{\bar{\alpha_t}}}.
    \label{eq:z0}
\end{equation}
Furthermore, given that $z^t_0$  has a bias that increases with larger $t$ , we further refine  $z^t_0$  by interpolating between the original image’s latent  $z_0$  and the predicted  $z^t_0$, as follows:
\begin{equation}
z^t_{in}=\sqrt{1-\bar{\alpha}_t} z_0+\left(1-\sqrt{1-\bar{\alpha}_t}\right) z^t_0,
\label{eq:xin}
\end{equation}
where $\sqrt{1 - \bar{\alpha}_t}$ decreases as  $t$  decreases, allowing  $z_0$  to take on a higher weight at larger  $t$  values, making the sample input $x^t_{in} = \mathcal{D}(z^t_{in})$ to the classifier progressively cleaner, enhancing reward accuracy at each denoising step. Additionally, inspired by the trajectory-level momentum update method, we propose a step-level momentum accumulation:
\begin{equation}
    g_{st} = \nabla_{z_t} R_a(f_\phi(x^t_{{in}}), y),  m^t_{st} = m^{t+1}_{st} + \frac{g_{st}}{\|g_{st}\|_1},
\label{eq:xin2}
\end{equation}
where  $g_{st}$ and $m^t_{st}$  denotes step-level gradient and momentum. We replace $\nabla_{z_t} R_a(f_\phi(\mathcal{D}(z_t)), y)$  in Eq.~\ref{eq:classgudied} with  $\text{sgn}(m^t_{st})$. Finally, by combining trajectory-level and step-level dual-path attack guidance, the generated images are fully aligned with attack effectiveness preference.

% \end{equation}

\noindent \textbf{Diffusion Augmentation.} 
Our dual-path attack optimization is based on direct backpropagation with a differentiable reward. Studies on HPA~\cite{liu2024alignment} have shown that direct backpropagation often leads to the diffusion model over-optimizing for the reward model. Similarly, in APA, this causes overfitting to the substitute classifier, limiting transfer attack performance. To address this, we propose diffusion augmentation which uses step-level outputs as data augmentation to enhance the generalization of the trajectory-level gradient $g_{tr}$. Specifically, we collect the step-level  $z^t_0$  generated during the denoising using Eq.~\ref{eq:z0}, and mix them with the trajectory-level final output $\bar{z}_0$:
\begin{equation}
    x^t_0= \varrho((\mathcal{D}({z^t_0})+\mathcal{D}({\bar{z}_0}))/2),
    \label{eq:mix}
\end{equation}
where $\varrho(\cdot)$ denotes differentiable data augmentation including random padding, resizing, and brightness adjustment. Appendix further shows that stronger data transformations (e.g., \cite{wang2023sit} used in $L_p$ attacks) can further boost performance, underscoring the scalability of our method. Finally, the trajectory-level gradient $g_{tr}$ in Eq.~\ref{eq:tr-level}  is enhanced to $g_{tr} = \nabla_{z_T} \frac{1}{T} \sum^{T}_{t=0} R_a(f_\phi(x^t_0), y)$.

Overall, we collectively refer to step-level attack guidance and diffusion augmentation as the attack guidance denoise process, as shown in Figure~\ref{fig:main}. To balance time efficiency and image quality, we apply attack guidance denoise only in the final  $T_a$  steps. We introduce APA-SG for skip gradient and APA-GC for gradient checkpointing.

\section{Experiments}
\vspace{-5pt}
\subsection{Experimental Settings }
\vspace{-5pt}
\textbf{Datasets and Models.} We choose the widely used ImageNet-compatible Dataset~\cite{kurakin2018adversarial}, consisting of 1,000 images from ImageNet’s validation set~\cite{imagenet}. Following~\cite{aca}, we select 6 convolutional neural networks (CNNs) and 4 vision transformers (ViTs) as target models for the attack.

\noindent \textbf{Attack Methods.} We compare with unrestricted attacks including SAE~\cite{sae}, cAdv~\cite{cadv}, tAdv~\cite{cadv}, ColorFool~\cite{colorfool}, and NCF~\cite{ncf}, as well as diffusion-based methods ACA~\cite{aca} and DiffPGD~\cite{diffpgd} in Table~\ref{tab:main}. Following~\cite{aca}, we use attack success rate (ASR, \%), the percentage of misclassified images—as the evaluation metric, reporting both white-box and black-box ASR.

\noindent \textbf{Implementation Details. \label{sec:imp}} We set attack guidance step  $T_a = 10$ , attack iterations  $N = 10$ , attack scale  $\epsilon_a = 0.4$ , and attack step size  $\mu = 0.04$. APA-SG adopts the entire inversion step of $T = 50$. APA-GC adopts $T = 10$ to improve efficiency. Our work is based on Stable Diffusion V1.5~\cite{ldm}. More implementation details in Appendix.

% \vspace{-5pt}

\renewcommand{\arraystretch}{0.75}
\begin{table*}[t]
\small
\centering
\caption{Attack performance comparison on normally trained CNNs and ViTs. We report attack success rates ASR (\%) of each method (“*” means white-box ASR), Avg. ASR refers to the average attack success rate on non-substitute models (black-box ASR).}
\resizebox{0.95\textwidth}{!}{
\begin{tabular}{c|cccccccccccc}
\toprule[1pt]
\multirow{6}{*}{\begin{tabular}[c]{@{}c@{}}Substitute\\Model\end{tabular}} & \multirow{6}{*}{Attack} & \multicolumn{10}{c}{Models} & \multirow{6}{*}{\makecell[c]{Avg.\\ASR (\%)}} \\ \cmidrule(lr){3-12}
 &  & \multicolumn{6}{c}{CNNs} & \multicolumn{4}{c}{Transformers} &  \\ \cmidrule(lr){3-8} \cmidrule(lr){9-12}
  &  & \makecell[c]{MN-v2 \\ \cite{mobilenetv2}} & \makecell[c]{Inc-v3 \\ \cite{inceptionv3}} & \makecell[c]{RN-50 \\ \cite{resnet}} & \makecell[c]{Dense-161\\ \cite{densenet}} & \makecell[c]{RN-152\\ \cite{resnet}} & \makecell[c]{EF-b7 \\ \cite{efficientnet}} & \makecell[c]{MobViT-s\\ \cite{mobvit}} & \makecell[c]{ViT-B\\ \cite{vit}} & \makecell[c]{Swin-B\\ \cite{swint}} & \makecell[c]{PVT-v2\\ \cite{pvtv2}} &  \\ \midrule[1.2pt]
\multirow{1}{*}{-} & Clean & 12.1 & 4.8 & 7.0 & 6.3 & 5.6 & 8.7 & 7.8 & 8.9 & 3.5 & 3.6 & 6.83 \\
 \midrule[1.2pt]
\multirow{11}{*}{MobViT-s} & SAE & 60.2 & 21.2 & 54.6 & 42.7 & 44.9 & 30.2 & 82.5* & 38.6 & 21.1 & 20.2 & 37.08 \\
 & cAdv & 41.9 & 25.4 & 33.2 & 31.2 & 28.2 & 34.7 & 84.3* & 32.6 & 22.7 & 22.0 & 30.21 \\
 & tAdv & 33.6 & 18.8 & 22.1 & 18.7 & 18.7 & 15.8 & 97.4* & 15.3 & 11.2 & 13.7 & 18.66 \\
 & ColorFool & 47.1 & 12.0 & 40.0 & 28.1 & 30.7 & 19.3 & 81.7* & 24.3 & 9.7 & 10.0 & 24.58 \\
 & NCF & {67.7} & 31.2 & 60.3 & 41.8 & 52.2 & 32.2 & 74.5* & 39.1 & 20.8 & 23.1 & 40.93 \\ 
 % \cmidrule(l){2-13}  
  & DiffPGD-MI & 59.9 & 42.4 & 48.1 & 44.6 & 38.5 & 38.1 & 95.7* & 29.0 & 24.9 & 42.5 & 40.89  \\
 & ACA & 66.2 & 56.6 & {60.6} & {58.1} & {55.9} & {55.5} & 89.8* & {51.4} & \underline{52.7} & {55.1} & {56.90} \\
 \cmidrule(l){2-13}
 % \cline{2-13}

 &\cellcolor{gray!20} \textbf{APA-SG(Ours)}& \cellcolor{gray!20}\underline{81.3} & \cellcolor{gray!20}\underline{66.3} & \cellcolor{gray!20}\underline{73.8} & \cellcolor{gray!20}\underline{71.5} & \cellcolor{gray!20}\underline{68.9} & \cellcolor{gray!20}\underline{65.0} & \cellcolor{gray!20}\underline{98.1*} & \cellcolor{gray!20}\underline{51.6} & \cellcolor{gray!20}{46.1} & \cellcolor{gray!20}\underline{68.2} & \cellcolor{gray!20}\underline{65.85} \\

 & \cellcolor{gray!20}\textbf{APA-GC(Ours)} & \cellcolor{gray!20}\textbf{88.3} & \cellcolor{gray!20}\textbf{77.1} & \cellcolor{gray!20}\textbf{86.6} & \cellcolor{gray!20}\textbf{81.2} & \cellcolor{gray!20}\textbf{81.2} & \cellcolor{gray!20}\textbf{78.4} & \cellcolor{gray!20}\textbf{99.4*} & \cellcolor{gray!20}\textbf{59.3} & \cellcolor{gray!20}\textbf{61.9} & \cellcolor{gray!20}\textbf{83.4} & \cellcolor{gray!20}\textbf{77.48} \\
\midrule[1.2pt]
\multirow{11}{*}{MN-v2} & SAE & 90.8* & 22.5 & 53.2 & 38.0 & 41.9 & 26.9 & 44.6 & 33.6 & 16.8 & 18.3 & 32.87 \\
 & cAdv & 96.6* & 26.8 & 39.6 & 33.9 & 29.9 & 32.7 & 41.9 & 33.1 & 20.6 & 19.7 & 30.91 \\
 & tAdv & 99.9* & 27.2 & 31.5 & 24.3 & 24.5 & 22.4 & 40.5 & 16.1 & 15.9 & 15.1 & 24.17 \\
 & ColorFool & 93.3* & 9.5 & 25.7 & 15.3 & 15.4 & 13.4 & 15.7 & 14.2 & 5.9 & 6.4 & 13.50 \\
 & NCF & 93.2* & 33.6 & {65.9} & 43.5 & {56.3} & 33.0 & 52.6 & 35.8 & 21.2 & 20.6 & 40.28 \\ 
 % \cmidrule(l){2-13}  
  & DiffPGD-MI & 97.4* & 54.1 & 68.2 & 57.8 & 56.6 & 52.1 & 68.0 & 28.7 & 22.9 & 41.8 & 50.02 \\
 & ACA & 93.1* & {56.8} & 62.6 & {55.7} & 56.0 & {51.0} & {59.6} & {48.7} & \underline{48.6} & {50.4} & {54.38} \\
 \cmidrule(l){2-13}

 & \cellcolor{gray!20} \textbf{APA-SG(Ours)} & \cellcolor{gray!20}\underline{99.8*} & \cellcolor{gray!20}\underline{80.4} & \cellcolor{gray!20}\underline{88.1} & \cellcolor{gray!20}\underline{83.0} & \cellcolor{gray!20}\underline{81.7} & \cellcolor{gray!20}\underline{78.8} & \cellcolor{gray!20}\underline{78.5} & \cellcolor{gray!20}\underline{55.9} & \cellcolor{gray!20}{39.5} & \cellcolor{gray!20}\underline{63.4} & \cellcolor{gray!20}\underline{72.14} \\

 & \cellcolor{gray!20}\textbf{APA-GC(Ours)} & \cellcolor{gray!20}\textbf{100*} & \cellcolor{gray!20}\textbf{91.4} & \cellcolor{gray!20}\textbf{97.7} & \cellcolor{gray!20}\textbf{95.5} & \cellcolor{gray!20}\textbf{95.0} & \cellcolor{gray!20}\textbf{91.8} & \cellcolor{gray!20}\textbf{93.2} & \cellcolor{gray!20}\textbf{74.3} & \cellcolor{gray!20}\textbf{59.0} & \cellcolor{gray!20}\textbf{85.2} & \cellcolor{gray!20}\textbf{87.01} \\

 \midrule[1.2pt]
\multirow{11}{*}{RN-50} & SAE & 63.2 & 25.9 & 88.0* & 41.9 & 46.5 & 28.8 & 45.9 & 35.3 & 20.3 & 19.6 & 36.38 \\
 & cAdv & 44.2 & 25.3 & 97.2* & 36.8 & 37.0 & 34.9 & 40.1 & 30.6 & 19.3 & 20.2 & 32.04 \\
 & tAdv & 43.4 & 27.0 & 99.0* & 28.8 & 30.2 & 21.6 & 35.9 & 16.5 & 15.2 & 15.1 & 25.97 \\

 & ColorFool & 41.6 & 9.8 & 90.1* & 18.6 & 21.0 & 15.4 & 20.4 & 15.4 & 5.9 & 6.8 & 17.21 \\
 & NCF & {71.2} & 33.6 & 91.4* & 48.5 & 60.5 & 32.4 & 52.6 & 36.8 & 19.8 & 21.7 & 41.90 \\ 
 % \cmidrule(l){2-13}  
  & DiffPGD-MI & 75.2 & 60.6 & 96.8* & 75 & 78.9 & 55.3 & 67.5 & 30.3 & 26.5 & 48.5 & 57.53 \\
 & ACA & 69.3 & {61.6} & 88.3* & {61.9} & {61.7} & {60.3} & {62.6} & {52.9} & \underline{51.9} & {53.2} & {59.49} \\
 \cmidrule(l){2-13}  
 & \cellcolor{gray!20} \textbf{APA-SG(Ours)} & \cellcolor{gray!20}\underline{89.0} & \cellcolor{gray!20}\underline{83.4} & \cellcolor{gray!20}\underline{99.6*} & \cellcolor{gray!20}\underline{89.6} & \cellcolor{gray!20}\underline{90.1} & \cellcolor{gray!20}\underline{77.3} & \cellcolor{gray!20}\underline{76.7} & \cellcolor{gray!20}\underline{58.5} & \cellcolor{gray!20} {45.7} & \cellcolor{gray!20}\underline{67.6} & \cellcolor{gray!20}\underline{75.32} \\
 & \cellcolor{gray!20}\textbf{APA-GC(Ours)} & \cellcolor{gray!20}\textbf{97.6} & \cellcolor{gray!20}\textbf{93.5} & \cellcolor{gray!20}\textbf{99.7*} & \cellcolor{gray!20}\textbf{97.6} & \cellcolor{gray!20}\textbf{98.4} & \cellcolor{gray!20}\textbf{91.1} & \cellcolor{gray!20}\textbf{90.9} & \cellcolor{gray!20}\textbf{75.6} & \cellcolor{gray!20}\textbf{63.8} & \cellcolor{gray!20}\textbf{83.7} & \cellcolor{gray!20}\textbf{88.02} \\

 \midrule[1.2pt]
\multirow{11}{*}{ViT-B} & SAE & 54.5 & 26.9 & 49.7 & 38.4 & 41.4 & 30.4 & 46.1 & 78.4* & 19.9 & 18.1 & 36.16 \\
 & cAdv & 31.4 & 27.0 & 26.1 & 22.5 & 19.9 & 26.1 & 32.9 & 96.5* & 18.4 & 16.9 & 24.58 \\
 & tAdv & 39.5 & 22.8 & 25.8 & 23.2 & 22.3 & 20.8 & 34.1 & 93.5* & 16.3 & 15.3 & 24.46 \\
 & ColorFool & 45.3 & 13.9 & 35.7 & 24.3 & 28.8 & 19.8 & 27.0 & 83.1* & 8.9 & 9.3 & 23.67 \\
 & NCF & 55.9 & 25.3 & 50.6 & 34.8 & 42.3 & 29.9 & 40.6 & 81.0* & 20.0 & 19.1 & 35.39 \\ 
 & DiffPGD-MI & 59.5 & 40.9 & 44.2 & 41.9 & 41.3 & 41.3 & 52.2 & 95.4* & 42.1 & 33.8 & 44.13  \\ 
 & ACA & {64.6} & {58.8} & {60.2} & {58.1} & {58.1} & {57.1} & {60.8} & 87.7* & {55.5} & {54.9} & {58.68} \\
 \cmidrule(l){2-13}

 & \cellcolor{gray!20} \textbf{APA-SG(Ours)} & \cellcolor{gray!20}\underline{69.3} & \cellcolor{gray!20}\underline{67.6} & \cellcolor{gray!20}\underline{67.5} & \cellcolor{gray!20}\underline{66.8} & \cellcolor{gray!20}\underline{65.4} & \cellcolor{gray!20}\underline{70.0} & \cellcolor{gray!20}\underline{67.6} & \cellcolor{gray!20}\textbf{99.2*} & \cellcolor{gray!20}\underline{62.5} & \cellcolor{gray!20}\underline{59.2} & \cellcolor{gray!20}\underline{66.21}\\
 % \rowcolor{gray!20}
 & \cellcolor{gray!20}\textbf{APA-GC(Ours)} & \cellcolor{gray!20}\textbf{77.0}&\cellcolor{gray!20}\textbf{74.8}&\cellcolor{gray!20}\textbf{75.4}&\cellcolor{gray!20}\textbf{75.9}&\cellcolor{gray!20}\textbf{75.4}&\cellcolor{gray!20}\textbf{76.8}&\cellcolor{gray!20}\textbf{74.5}&\cellcolor{gray!20}98.4*&\cellcolor{gray!20}\textbf{73.4}&\cellcolor{gray!20}\textbf{70.2}& \cellcolor{gray!20}\textbf{74.82}\\
 \bottomrule[1pt]
\end{tabular}}
% \vspace{-10pt}
\label{tab:main}
% \label{tab:attacknormally}
\end{table*}

\subsection{Attack Performance Comparison}
\vspace{-5pt}
To evaluate the performance of our APA framework, we select 10 models as target models, including both CNN and transformer architectures. The attack performance comparison is shown in Table~\ref{tab:main}.

\noindent \textbf{Non-diffusion-based Attacks.} Texture-based tAdv achieves a higher white-box ASR but lower black-box ASR than color-based methods such as NCF and SAE. NCF demonstrates the highest transferability, achieving an average ASR of 39.6\% across four models. Our APA, which modifies multiple input semantics simultaneously, surpasses single-semantic attacks, with notable improvements and 30.2\% (APA-SG) and 42.2\% (APA-GC) in black-box ASR across four models than NCF.

\noindent \textbf{Diffusion-based Attacks.} Since both our method and ACA use momentum-based settings, we upgrade DiffPGD to a momentum version, DiffPGD-MI, for comparison. DiffPGD-MI generates unrestricted adversarial examples by first applying  $L_p$ norm perturbations, and then performing diffusion-based image translation. ACA directly optimizes the input image’s latent space, generally achieving higher black-box transferability, especially across architectures. Our method incorporates dual-path attack guidance and diffusion augmentation, enabling APA-SG (with the same gradient backpropagation as ACA) to improve black-box ASR by 12.5\%, while APA-GC improves black-box performance by 24.4\% over ACA across four models.

\noindent \textbf{Overall.} Our method outperforms existing unrestricted adversarial attacks in terms of black-box transferability, whether within CNN or transformer architectures or in cross-architecture attacks. Additionally, due to its more precise gradient-guided attack, APA-GC achieves an average performance improvement of 11.9\% over APA-SG.

\begin{table*}[t]
\small
\centering
\caption{Attack performance on adversarial defense methods, ViT-B as the substitute model.}
\resizebox{0.95\textwidth}{!}{
\begin{tabular}{cccccccccccccc}
\toprule[1pt]
Attack & HGD & R\&P & NIPS-r3 & JPEG & \makecell[c]{Bit-\\Red} & DiffPure & \makecell[c]{Inc-\\v3$_{ens3}$} & \makecell[c]{Inc-\\v3$_{ens4}$} & \makecell[c]{IncRes-\\v2$_{ens}$} & Res-De & \makecell[c]{Shape-\\Res} & \makecell[c]{ViT-B-\\CvSt}  & Avg. ASR (\%) \\ \midrule[1pt]
Clean & 1.2 & 1.8 & 3.2 & 6.2 & 17.6 & 15.4 & 6.8 & 8.9 & 2.6 & 4.1 & 6.7 & 8.4 &  6.91 \\ \midrule[1pt]
SAE & 21.4 & 19.0 & 25.2 & 25.7 & 43.5 & 39.8 & 25.7 & 29.6 & 20.0 & 35.1 & 49.6 & 38.9 &  31.13 \\
cAdv & 12.2 & 14.0 & 17.7 & 11.1 & 33.9 & 32.9 & 19.9 & 23.2 & 14.6 & 16.2 & 25.3 & 20.6 &  20.13 \\
tAdv & 10.9 & 12.4 & 14.4 & 17.8 & 29.6 & 21.2 & 17.7 & 19.0 & 12.5 & 16.4 & 25.4 & 11.3 &  17.38 \\
ColorFool & 9.1 & 9.6 & 15.3 & 18.0 & 37.9 & 33.8 & 17.8 & 21.3 & 10.5 & 20.3 & 35.0 & 31.2 &  21.65 \\
NCF & 22.8 & 21.1 & 25.8 & 26.8 & 43.9 & 39.6 & 27.4 & 31.9 & 21.8 & 34.4 & 47.5 & 35.8 &  31.57 \\
DiffPGD-MI & 25.7 & 28.3 & 29.9 & 32.2 & 38.8 & 27.4 & 32.1 & 32.9 & 28.1 & 40.0 & 45.5 & 19.7 & 31.72 \\ 
ACA & {52.2} & {53.6} & {53.9} & {59.7} & {63.4} & {63.7} & {59.8} & {62.2} & {53.6} & {55.6} & {60.8} & \textbf{51.1}  & {57.47} \\ \midrule
\rowcolor{gray!20}
\textbf{APA-SG(Ours)} & \underline{61.5} & \underline{61.0}& \underline{63.8}  & \underline{66.7} & \underline{71.0} & \underline{63.0} & \underline{66.8} & \underline{67.4} & \underline{63.1} & \underline{64.5} & \underline{68.7} & \underline{45.6}  & \underline{63.59} \\ 
\rowcolor{gray!20}
\textbf{APA-GC(Ours)} & \textbf{73.5} & \textbf{71.1} & \textbf{72.4} & \textbf{72.4} & \textbf{74.3} & \textbf{71.2} & \textbf{72.5} & \textbf{73.6} & \textbf{71.4} & \textbf{72.4} & \textbf{75.5} & 42.1  & \textbf{70.20} \\ 
\bottomrule[1pt]
\end{tabular}}
\label{tab:attackdefense}
% \vspace{-5pt}

% \vspace{-15pt}
\label{tab:defense}
\vspace{-10pt}
\end{table*}

% \vspace{-5pt}
\subsection{Attacks on Adversarial Defense}\vspace{-5pt}To evaluate unrestricted adversarial attacks against existing defenses, we select adversarially trained models (Inc-v3${ens3}$, Inc-v3${ens4}$, Inc-v2${ens}$\cite{eat}, ViT-B-CvSt\cite{singh2023revisiting}) and preprocessing defenses (HGD~\cite{HGD}, R\&P~\cite{randp}, NIPS-r3, JPEG~\cite{jpeg}, Bit-Red~\cite{bitred}, DiffPure~\cite{diffpure}). Additionally, shape-texture debiased models (ResNet50-Debiased (Res-De)\cite{debaising}, Shape-ResNet (Shape-Res)\cite{towardstexture}) are selected to counter unrestricted adversarial examples, as shown in Table~\ref{tab:defense}. We use ViT-B as the substitute model and Inc-v3${ens3}$ as the target model for input preprocessing defenses. Since existing defenses mainly address $L_p$ attacks, they remain ineffective against unrestricted adversarial attacks. With an advanced APA framework, our APA-GC achieves a 12.7\% improvement on Avg. ASR over SOTA method.

% \vspace{-5pt}
\begin{table}
% \vspace{-20pt}
\centering
\caption{Quantitative comparison of image quality. VCA denotes only using visual consistency alignment. APA-GC-P denotes prompt-based optimization.}
\resizebox{0.46\textwidth}{!}{
\begin{tabular}{c|ccccc|c}
\toprule[1pt]
Attack     & LPIPS$\downarrow$         & SSIM$\uparrow$          &   \begin{tabular}[c]{@{}c@{}}CLIP \\ Score$\uparrow$\end{tabular}           & \begin{tabular}[c]{@{}c@{}}NIMA-\\ AVA$\uparrow$\end{tabular} &  \begin{tabular}[c]{@{}c@{}}CNN-\\IQA$\uparrow$\end{tabular}        & \begin{tabular}[c]{@{}c@{}}Avg.\\ ASR$\uparrow$\end{tabular} \\ \midrule
Clean      &  0.00 & 1.00 & 1.00 & 4.99 & 0.58 & 6.83 \\
\rowcolor{gray!20}
VCA       &  0.05 & 0.85 & 0.97 & 5.13  & 0.63 & 7.60
  \\ \midrule
SAE        & 0.43          & 0.79          & 0.81         & 5.00                                                         & 0.53          & 36.38                                                 \\
cAdv       & \underline{0.15}          & \textbf{0.98}          & \underline{0.88} & 4.85                                                     & 0.55          & 32.04                                                 \\
NCF        & 0.40          & \underline{0.83}          & 0.83         & 4.97                                                      & 0.57          & 41.90                 \\                              
DiffPGD-MI & 0.29          & 0.71          & 0.85          & 4.69                                                   & 0.61          & 57.53                                                 \\
ACA        & 0.37          & 0.61          & 0.79          & \underline{5.38}                                                 & \underline{0.65}          & 59.49                                                 \\ 
\midrule
\rowcolor{gray!20}
\textbf{APA-SG(Ours)}     & 0.25          & 0.67          & 0.86        & 5.29                                                         & 0.62          & \underline{75.32}                                                 \\
\rowcolor{gray!20}
\textbf{APA-GC(Ours)}   & 0.23          & 0.69          & 0.83         & \textbf{5.39}                                             & \textbf{0.67} & \textbf{88.02}                                                 \\  
\rowcolor{gray!20}
\textbf{APA-GC-P(Ours)}   & \textbf{0.09} & {0.82}          &  \textbf{0.91}         & 5.22                                                       & 0.63          & 62.08                                                 \\
 \bottomrule[1pt]
\end{tabular}}
\label{tab:vis}
% \vspace{-10pt}
\end{table}

\subsection{Visual Quality Comparison}
\vspace{-5pt}
We compare the visual performance of the top five attack methods based on attack performance in Table~\ref{tab:main}, using RN-50 as the substitute model.

\begin{figure*}[!t]
    \centering
    \resizebox{\textwidth}{!}{
    \includegraphics{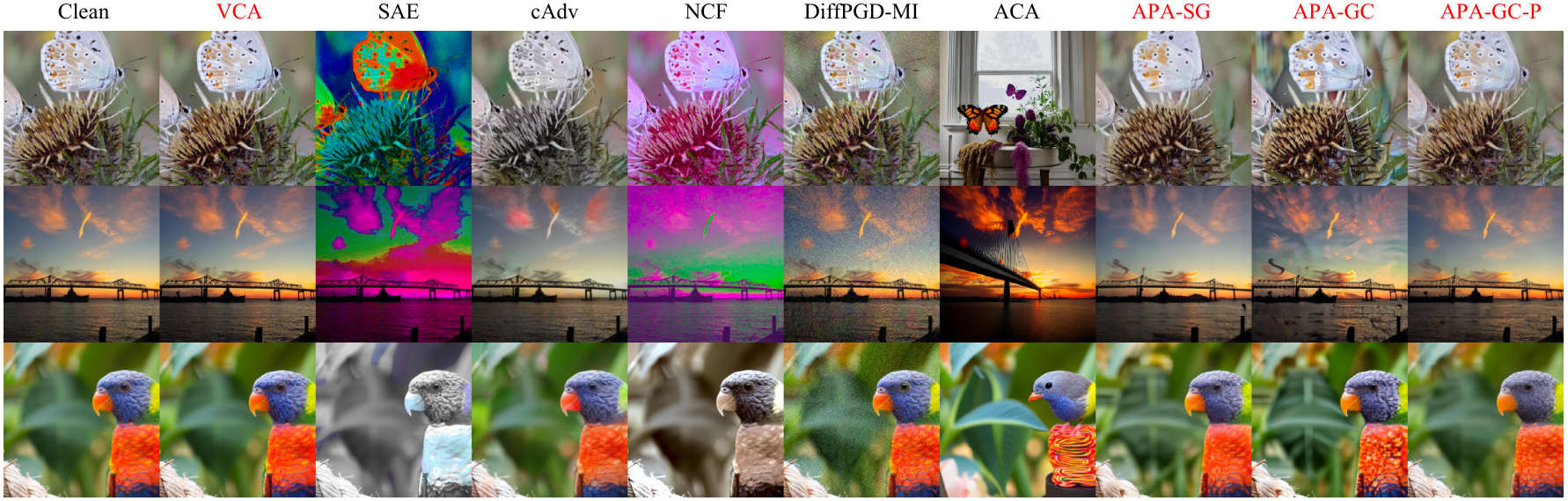}}
    \vspace{-20pt}
    \caption{Qualitative comparison of image quality.}
    \vspace{-10pt}
    \label{fig:vis}
\end{figure*}

\noindent \textbf{Quantitative Comparison.} We use reference-based metrics (LPIPS, SSIM, CLIP Score~\cite{clip}) to evaluate visual similarity in terms of distribution, structure, and semantics, and no-reference aesthetic metrics (NIMA-AVA~\cite{ava}, CNN-IQA~\cite{pyiqa}) to assess aesthetic quality, as shown in Table~\ref{tab:vis}. 1) Our visual consistency alignment (VCA) maintains strong visual consistency while achieving a higher aesthetic score compared to clean images. 2) Our method achieves higher aesthetic scores compared to non-diffusion-based attacks and DiffPGD owing to stable diffusion’s strong generative capabilities. 3) It also shows improved visual similarity metrics over ACA, primarily due to our visual consistency alignment. 4) Within our framework, APA-GC-P which optimizes prompt instead of latent (see Section~\ref{apa-p}) has the best visual consistency.

% \vspace{-5pt}
\noindent \textbf{Qualitative Comparison.} Due to the inability of quantitative metrics to fully measure visual consistency, we perform a qualitative analysis in Figure~\ref{fig:vis}.  SAE with NCF alters the original style, DiffPGD-MI introduces noticeable perturbations, and cAdv affects authenticity by changing colors. ACA disrupts the original structure. In contrast, our APA preserves structure and color, making only subtle, natural adjustments mainly in the background, resulting in a more visually consistent effect.

\subsection{Ablation Studies}
\vspace{-5pt}
The previous section has discussed the ablation of visual consistency alignment (Figure~\ref{fig:m1}) and gradient backpropagation (Table~\ref{tab:main}). Here, we focus on analyzing the remaining key modules and design. All experiments utilize RN-50 as the substitute model. Time analysis is discussed in Appendix.

\begin{figure}[!t]
    \centering
    \scalebox{0.5}{
    \includegraphics{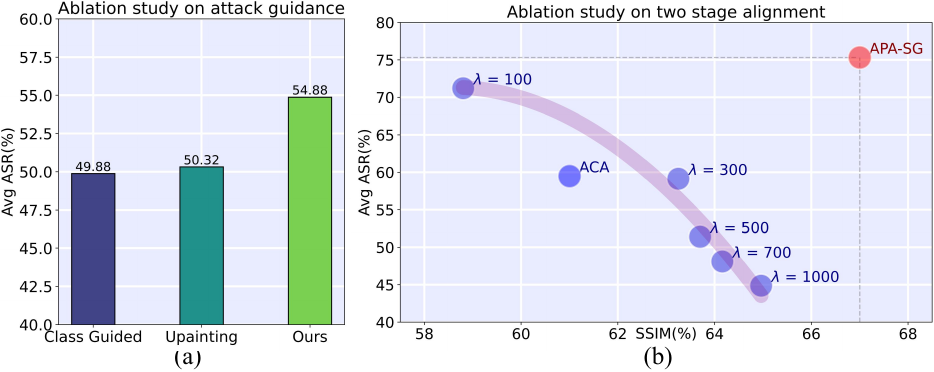}}
    % \vspace{-10pt}
    \caption{(a) shows the comparison with different optimization guidance. (b) Performance comparison of our two-stage APA-SG and one-stage alignment under $\lambda$-controlled visual consistency.}
    \vspace{-12pt}
    \label{fig:aba}
\end{figure}

\begin{table}
% \vspace{-20pt}
\centering
\caption{Ablation studies on key modules. L denotes latent-based optimization, P denotes prompt-based optimization.}
\scalebox{0.7}{
\begin{tabular}{cccc|cc}
\toprule[1pt]
\multirow{2}{*}{\begin{tabular}[c]{@{}c@{}}Optimized\\ Params\end{tabular}} & \multirow{2}{*}{\begin{tabular}[c]{@{}c@{}}Dual-path \\ Guidance\end{tabular}} & \multirow{2}{*}{\begin{tabular}[c]{@{}c@{}}Diffusion \\ Augmentation\end{tabular}} & \multirow{2}{*}{\begin{tabular}[c]{@{}c@{}}Backpro-\\pagation\end{tabular}} & \multirow{2}{*}{\begin{tabular}[c]{@{}c@{}}White-box\\ ASR(\%)\end{tabular}}& \multirow{2}{*}{\begin{tabular}[c]{@{}c@{}}Black-box\\ ASR(\%)\end{tabular}} \\
 &  &  &  &  \\ \midrule
L &  &  & SG & 96.8 & 48.28 \\
L & \checkmark &  & SG & \textbf{99.7} & 54.88 \\
L &  & \checkmark & SG & 92.1& 62.38 \\
L & \checkmark & \checkmark & SG & \underline{99.6} & \underline{75.32} \\
L & \checkmark & \checkmark & GC & \textbf{99.7} & \textbf{88.02} \\
P & \checkmark & \checkmark & GC & 99.5 & 62.08 \\ \bottomrule[1pt]
\end{tabular}}
\label{tab:aba}
    
  \vspace{-10pt}
\end{table}
\noindent \textbf{Key Modules.} Rows 1 and 2 of Table~\ref{tab:aba} show that the dual-path attack guidance module improves black-box attack performance by 6.6\% compared to only trajectory-level guidance. To further validate the superiority of our attack guidance, we re-implement class-guided~\cite{guide-diffusion} and Upainting~\cite{li2022upainting} by applying $R_a$ for adversary preferences alignment. Figure~\ref{fig:aba}(a) shows improved attack performance with our method, which benefits from more accurate attack reward guidance through clear $z^t_{in}$ in Eq.~\ref{eq:xin} and step-level momentum accumulation in Eq.~\ref{eq:xin2}. Rows 1 and 3 of Table~\ref{tab:aba} demonstrate that diffusion augmentation mitigates the limitations of direct backpropagation overfitting to the substitute model, improving black-box performance by 14.1\% with only a slight decrease in white-box performance. Rows 3 and 4 in Table~\ref{tab:aba} show that diffusion augmentation combined with dual-path attack guidance effectively improves black-box attack performance.

\noindent \textbf{Two-stage vs. One-stage Alignment.} To validate the advantages of our two-stage alignment, we adapt APA into a single-stage alignment (APA*): replacing LoRA-based visual alignment and incorporating joint optimization in the second stage, i.e., $R_a = R_a - \lambda \|z_0 - \bar{z}_0\|_2$. Experimental results in Figure~\ref{fig:aba}(b) demonstrate: 1) One-stage alignment (both APA* and ACA) suffer from reward hacking due to conflicting objectives during joint optimization (as $\lambda$ increases, Avg. ASR decreases while SSIM increases). 2) Our two-stage APA maximizes attack performance within the optimal solution space of visual consistency, achieving closer Pareto optimality.

\noindent \textbf{Optimized Parameters. \label{apa-p}} Row 6 in Table~\ref{tab:aba} shows the attack performance with prompt-based optimization (APA-GC-P), which optimizes text features $\tau_\theta(c)$ with gradient checkpointing. Compared to the direct correspondence between the latent and image spaces, prompt-based optimization indirectly guides image generation through $\epsilon_{\theta+\Delta \theta}(z_t, t, c)$, resulting in lower attack efficacy than APA-GC. However, as shown in Table~\ref{tab:vis}, APA-GC-P demonstrates improved visual consistency, offering attackers greater flexibility depending on the application scenario.

\noindent \textbf{Scalability.} To demonstrate the flexibility and scalability of our framework, we extend it to various diffusion models (e.g., ControlNet~\cite{li2025controlnet}) in Appendix and different tasks including targeted attacks, visual question answering, and object detection in Appendix.

\subsection{Hyper-parameters tuning \label{sec:hyper}}
\begin{figure}[!t]
    \centering
    \scalebox{0.26}{
    \includegraphics{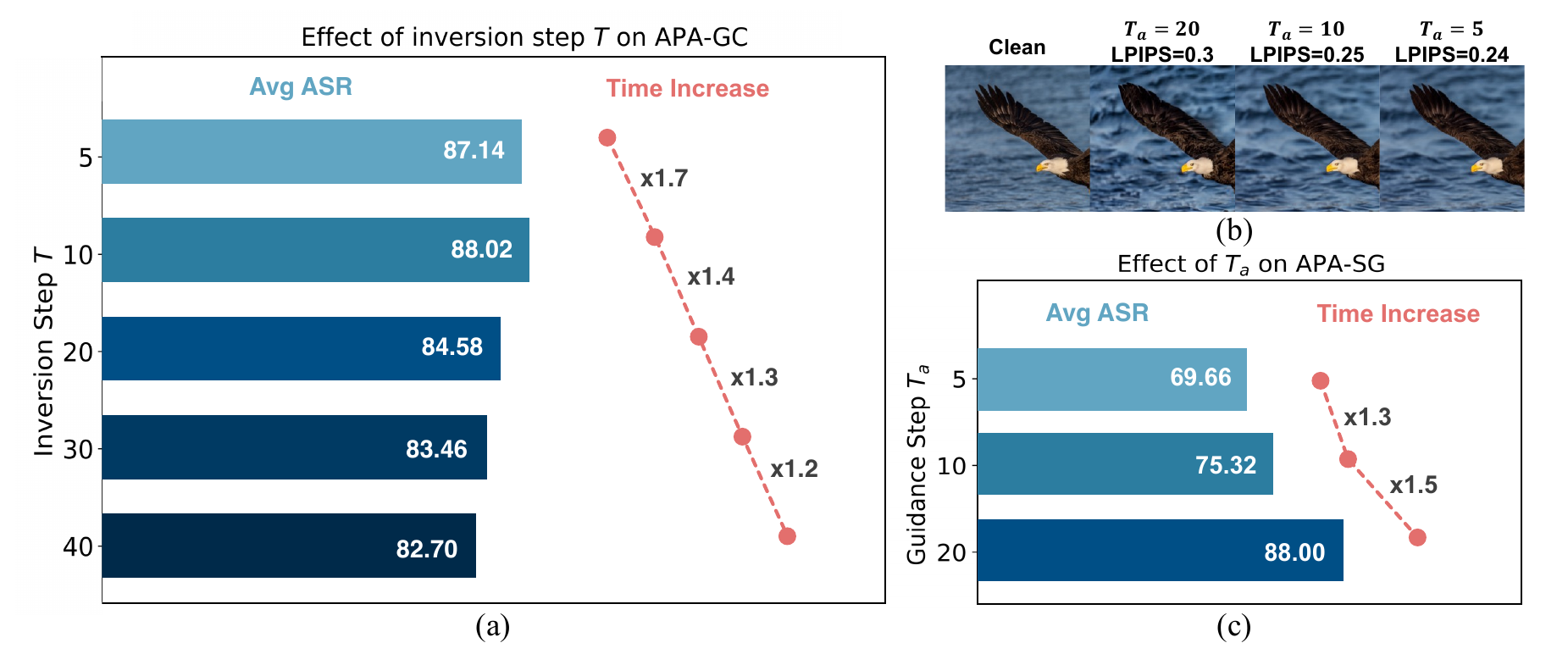}}
    % \vspace{-10pt}
    \caption{Hyper-parameters tuning on $T$ and $T_a$. RN-50 as the substitute model.}
    % \vspace{-20pt}
    \label{fig:tuning}
\end{figure}

\vspace{-5pt}
\textbf{Guidance Step $T_a$.} Figure~\ref{fig:tuning} (b), (c) show the impact of $T_a$ on performance. As $T_a$ increases, attack performance improves, time cost rises, and image quality deteriorates. Considering these factors, we choose $T_a = 10$.

\noindent \textbf{Inversion Step $T$.} APA-GC employs gradient checkpointing to save memory at the cost of additional time. To improve efficiency, we investigate the impact of reducing inversion steps $T$ on performance. Figure~\ref{fig:tuning}(a) shows that setting $T$ below $T_a$ reduces attack performance due to insufficient guidance, while exceeding $T_a$ also degrades attack performance due to bias introduced by overly deep gradient chains. Thus, we set $T = T_a = 10$.

\section{Conclusion}
% \vspace{-10pt}
In this paper, we broaden the application of preference alignment, reformulating unrestricted adversarial example generation as an adversary preferences alignment problem. However, the inherently conflicting objectives of visual consistency and attack effectiveness significantly increase the difficulty of alignment. To address this challenge, we propose Adversary Preferences Alignment (APA), a two-stage framework that first establishes visual consistency through LoRA-based alignment guided by a rule-based similarity reward, and then enhances attack effectiveness via dual-path attack guidance and diffusion-based augmentation. Experimental results demonstrate that APA achieves superior black-box transferability while preserving high visual consistency. We hope our work serves as a bridge between preference alignment and adversarial attacks, and inspires further research on adversarial robustness from an alignment perspective.
{
    \small
    \bibliographystyle{ieeenat_fullname}
    \bibliography{ref}
}

\end{document}